\documentclass[conference]{IEEEtran}
\IEEEoverridecommandlockouts
\usepackage{cite}
\usepackage{amsmath,amssymb,amsfonts}
\usepackage{algorithmic}
\usepackage{graphicx}
\usepackage{textcomp}
\usepackage{caption}
\usepackage{multirow}

\usepackage[ruled,linesnumbered]{algorithm2e}
\usepackage{xcolor}
\def\BibTeX{{\rm B\kern-.05em{\sc i\kern-.025em b}\kern-.08em
    T\kern-.1667em\lower.7ex\hbox{E}\kern-.125emX}}
\begin{document}

\title{Adapting LLMs for Efficient Context Processing through Soft Prompt Compression\\
}

\author{Cangqing Wang$^1$, Yutian Yang$^1$\\ \qquad \ Ruisi Li$^2$, Dan Sun$^2$, Ruicong Cai$^2$, Yuzhu Zhang$^2$, Chengqian Fu$^3$ and Lillian Floyd$^*$
\thanks{$^1$Cangqing Wang be with Boston University, MA 02215, USA {\tt\small \{kriswang\}@bu.edu}}
\thanks{$^1$Yutian Yang be with University of California, Davis, CA 95616, USA {\tt\small \{yytyang\}@ucdavis.edu}}
\thanks{$^2$Ruisi Li is an independent researcher. Correspondence to Ruisi Li via email: {\tt\small \{irisli7728\}@gmail.com}}
\thanks{$^2$Dan Sun be with Washington University in St. Louis, MO 63130, USA 
{\tt\small \{sun.dan\}@wustl.edu}}
\thanks{$^2$Ruicong Cai is an independent researcher. Correspondence to Ruicong Cai via email: {\tt\small \{aruiqian\}@gmail.com}}
\thanks{$^2$Yuzhu Zhang be with Carnegie Mellon University, Pittsburgh, PA 15213 {\tt\small \{yuzhuz\}@alumni.cmu.edu}}
\thanks{$^3$Chengqian Fu is an independent researcher. Correspondence to Chengqian Fu via email: {\tt\small \{cqfu728\}@gmail.com}}
\thanks{$^*$Lilian Floyd be with Georgia Institue of Technology, GA 30332, USA {\tt\small \{lfloyd74\}@gatech.edu}}
}

\maketitle

\begin{abstract}
The rapid advancement of Large Language Models (LLMs) has inaugurated a transformative epoch in natural language processing, fostering unprecedented proficiency in text generation, comprehension, and contextual scrutiny. Nevertheless, effectively handling extensive contexts, crucial for myriad applications, poses a formidable obstacle owing to the intrinsic constraints of the models' context window sizes and the computational burdens entailed by their operations. This investigation presents an innovative framework that strategically tailors LLMs for streamlined context processing by harnessing the synergies among natural language summarization, soft prompt compression, and augmented utility preservation mechanisms. Our methodology, dubbed SoftPromptComp, amalgamates natural language prompts extracted from summarization methodologies with dynamically generated soft prompts to forge a concise yet semantically robust depiction of protracted contexts. This depiction undergoes further refinement via a weighting mechanism optimizing information retention and utility for subsequent tasks. We substantiate that our framework markedly diminishes computational overhead and enhances LLMs' efficacy across various benchmarks, while upholding or even augmenting the caliber of the produced content. By amalgamating soft prompt compression with sophisticated summarization, SoftPromptComp confronts the dual challenges of managing lengthy contexts and ensuring model scalability. Our findings point towards a propitious trajectory for augmenting LLMs' applicability and efficiency, rendering them more versatile and pragmatic for real-world applications. This research enriches the ongoing discourse on optimizing language models, providing insights into the potency of soft prompts and summarization techniques as pivotal instruments for the forthcoming generation of NLP solutions.
\end{abstract}

\begin{IEEEkeywords}
Knowledge Graph Reasoning, Reinforcement Learning, Reward Shaping, Transfer Learning
\end{IEEEkeywords}

\section{Introduction}
The finite context window size inherent in most LLMs constrains their ability to fully grasp and utilize extensive textual information, thereby limiting their efficacy on tasks demanding profound comprehension of lengthy documents. Additionally, the substantial computational resources requisite for LLM processing present another obstacle, particularly for applications necessitating swift responsiveness and high throughput. These challenges underscore the imperative for innovative methodologies aimed at enhancing LLM efficiency and context management capabilities without compromising performance.

This paper introduces a pioneering framework, titled Soft Prompt Compression for LLMs (SPC-LLM), which aims to overcome these constraints by amalgamating the principles of soft prompt compression with natural language summarization techniques. 
the integration of soft prompt compression with LLMs could benefit from references to pioneering studies on prompt engineering or soft prompts in NLP\cite{Su_2022}. Our approach endeavors to tailor LLMs for streamlined context processing, enabling them to navigate extensive textual data more adeptly while alleviating computational burdens.

Our strategy comprises two primary facets: firstly, leveraging natural language summarization to distill protracted texts into succinct, content-rich summaries, and secondly, integrating these summaries into the model's input via trainable soft prompts. This dual-pronged approach extends the effective context window of LLMs and fosters a nuanced comprehension and generation of text predicated on diverse information reservoirs. By condensing the context into a compact, information-dense format, SPC-LLM substantially diminishes computational overheads, rendering the deployment of LLMs more viable across a broad array of applications.

We delineate a comprehensive methodology for implementing soft prompt compression alongside natural language summarization within LLMs, elucidating how this amalgamation augments model performance on tasks necessitating comprehension of extended contexts. Moreover, we furnish empirical substantiation from a series of experiments evincing the efficacy of SPC-LLM in enhancing the efficiency and precision of LLMs across various NLP tasks.

The structure of this paper is as follows: we commence with a review of pertinent literature concerning LLM efficiency and context processing. Subsequently, we expound upon the proposed SPC-LLM framework, delineating its constituents and the integration process. We subsequently expound upon the experimental setup and present our findings, highlighting the advantages of our approach. Finally, we deliberate on the implications of our work and posit avenues for future research endeavors in this domain.

\section{Prior Work}
The endeavor to augment the efficiency and contextual processing capabilities of Large Language Models (LLMs) stands as a focal point in contemporary natural language processing (NLP) inquiry. This section provides an overview of notable advancements in three pertinent domains: traditional methodologies for managing lengthy contexts within LLMs, the genesis and utilization of soft prompts, and innovations in text summarization techniques aimed at contextual information compression.

\begin{figure}[htbp]
\centerline{\includegraphics[height=6cm]{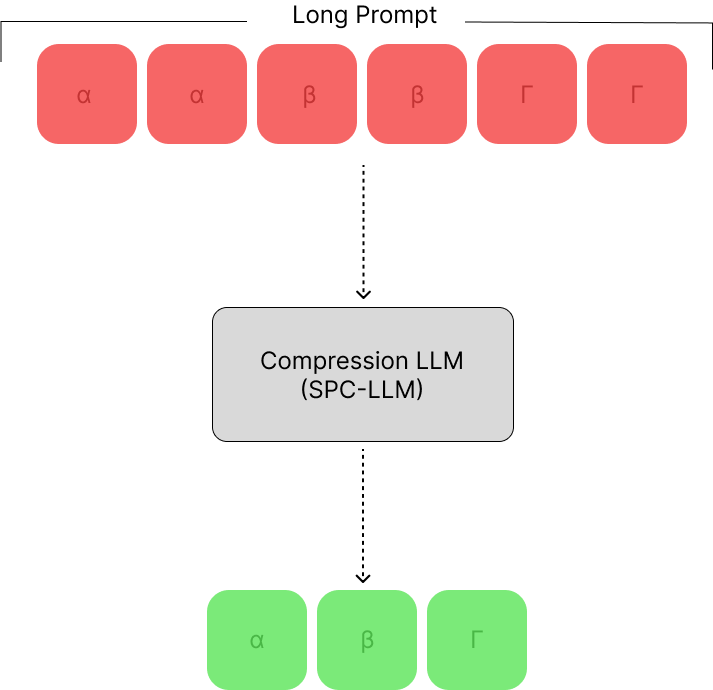}}
\caption{An example of successful prompt compression with SPC. The compressed prompt (green) in order to obtain a shorter length and maintain transferability and utility simultaneously than the original long prompt (red).}
\label{fig-1}
\end{figure}

\begin{figure}[htbp]
\centerline{\includegraphics[height=4cm]{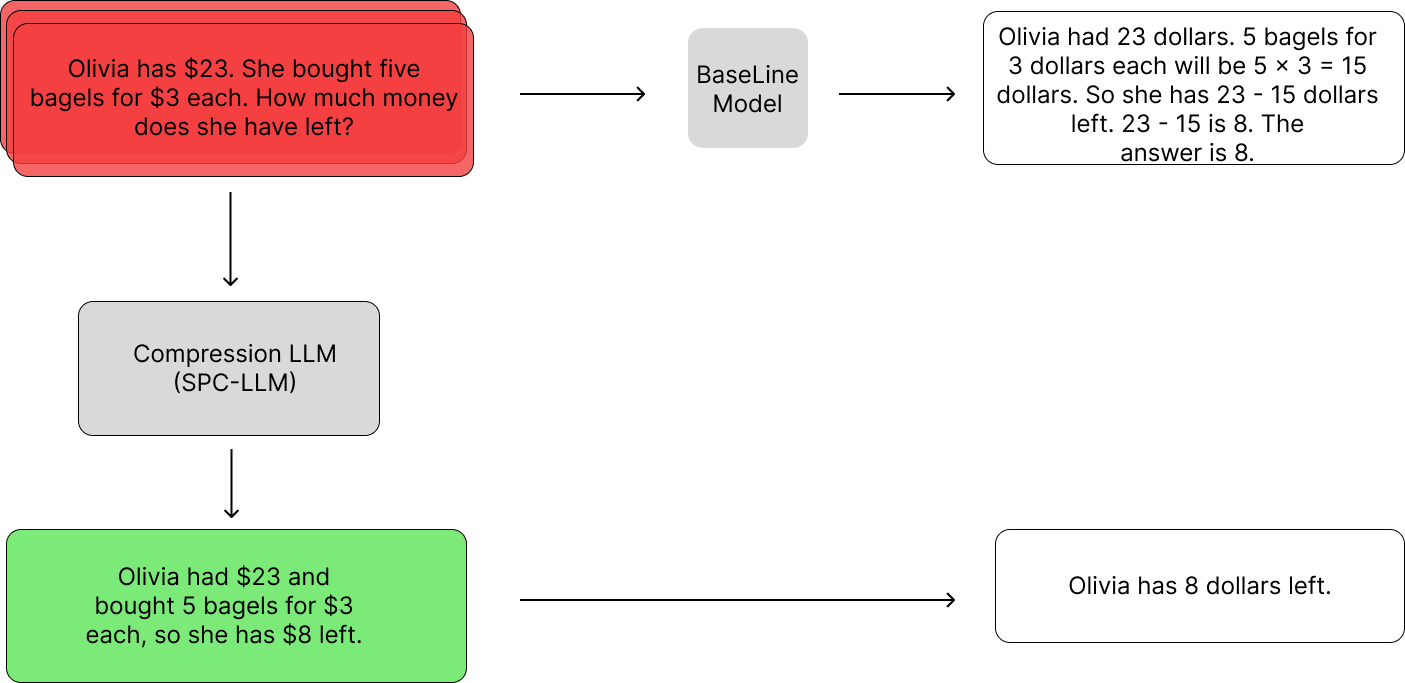}}
\caption{The illustration of SPC shows the compressed conversational answer expect with question.}
\label{fig-2}
\end{figure}
\subsection{Addressing Lengthy Contexts in LLMs}
Initial efforts to expand the contextual scope of LLMs primarily concentrated on architectural innovations. Transformer-based models such as Longformer \cite{beltagy2020longformer} and BigBird \cite{zaheer2021big} introduced alterations to the conventional attention mechanism, facilitating the efficient analysis of extended textual sequences. These models employ sparse attention patterns and memory-efficient algorithms to mitigate computational overheads associated with handling extensive contexts. Nevertheless, notwithstanding their augmented capacity, these architectural refinements frequently necessitate significant alterations to pre-existing models and entail substantial retraining endeavors.

\subsection{Soft Prompts in NLP}
The notion of soft prompts, first introduced by Li and Liang \cite{li2021prefixtuning} and further elucidated by Lester \cite{lester2021power}, signifies a departure towards more adaptable and parameter-efficient approaches for tailoring LLMs to specific tasks. Unlike rigidly defined textual prompts, soft prompts comprise trainable embeddings optimized alongside the model to enhance performance on targeted tasks. This adaptability facilitates effective fine-tuning of LLMs sans exhaustive retraining, thereby enabling their adaptation across diverse applications. Nonetheless, the amalgamation of soft prompts with the objective of enhancing context compression and efficiency within LLMs remains a relatively underexplored domain.

\subsection{Text Summarization for Contextual Compression}
Text summarization methodologies, encompassing both extractive and abstractive approaches, have been harnessed to condense protracted documents into succinct representations while retaining salient information. Advancements in summarization techniques, exemplified by models like BART \cite{lewis2019bart}, have exhibited considerable promise in generating coherent and concise summaries. Although these models adeptly distill lengthy texts, their direct utilization as a preprocessing step for LLMs in tasks necessitating nuanced comprehension of extended contexts has not been extensively investigated.

The fusion of soft prompts with summarization techniques to augment the processing of lengthy contexts in LLMs epitomizes a novel convergence of these research trajectories. Our study builds upon these foundational principles, proposing a unified framework that capitalizes on the strengths of soft prompts and advanced summarization methodologies to alleviate the constraints of existing LLMs in efficiently handling extensive textual information.

\section{Methodology}
Integrating summary vectors with natural language formatted prompts, the fusion of utility preservation with information compression, and the conceptualization of soft prompts are explored within a comprehensive mathematical model.
In this section, our endeavor lies in the construction of a holistic mathematical framework, embodying the amalgamation of summary vectors with prompts formatted in natural language, the convergence of utility retention with information condensation, and the notion of soft prompts. Let us consider the following components:

\begin{itemize}
    \item An original document $D$
    \item A function $f_{NL}$ for generating prompts in natural language, yielding $P_{NL}$
    \item A function $f_{S}$ for the generation of summary vectors, yielding $V_{S}$
    \item A parameter matrix $S$ for soft prompt
    \item A predictive function of the language model, denoted as  $f_{LM}$
\end{itemize}

The expression of the integrated model we aspire to formulate is as follows:
\begin{equation}
    C = f_{LM}(S \oplus f_{S}(f_{NL}(D)))
\end{equation}

Where:
\begin{itemize}
    \item $f_{NL}(D)$ denotes the conversion of the original document $D$ into the prompt formatted in natural language, denoted as $P_{NL}$.
    \item $f_{S}(P_{NL})$ signifies the further compression of $P_{NL}$ into the summary vector $V_{S}$.
    \item $S$ represents the parameter matrix of trained soft prompts, meticulously optimized to augment the model's versatility across distinct tasks.
    \item $\oplus$ denotes the operation of concatenation, amalgamating the soft prompt vector with the summary vector $V_{S}$.
    \item $f_{LM}$ represents the predictive capability of the language model, which takes the concatenated vector as input and generates the ultimate compressed representation $C$.
\end{itemize}

In this expression:
\begin{itemize}
    \item Utility preservation is attained through the strategic design of functions $f_{S}$ and $f_{NL}$ to uphold the maximum preservation of information.
    \item The compression of information is executed via function$f_{S}$, which compresses the NL formatted prompt $P_{NL}$ into the summary vector $V_{S}$.
    \item The soft prompt is optimized during the training process to learn how to most effectively utilize the compressed information $V_{S}$ to improve the model’s performance on specific tasks.

\end{itemize}
Utilizing this methodology, we harness the robust information compression potential inherent in NL formatted prompts and summary vectors. Additionally, we bolster the adaptability and efficacy of the model for particular downstream tasks by integrating soft prompts, thereby amplifying its capacity for generalization \cite{li2024enhancing}.

\section{Experiments}
The principal aim of these experiments is to assess the efficacy of our proposed methodology in enhancing the efficiency and performance of Large Language Models (LLMs) during the processing of expansive textual contexts. Our specific focus lies in evaluating:
The efficacy of condensed contextual representations produced via natural language summarization and the influence of integrating soft prompt compression on the efficacy of LLMs across diverse NLP tasks.

To achieve this aim, we meticulously designed our experiments to measure two critical aspects of our methodology's impact on LLMs. First, we evaluated the quality and utility of the condensed contextual representations generated by our advanced natural language summarization techniques. This assessment focused on determining how effectively our approach could distill essential information from extensive textual data, thereby facilitating a more efficient processing by LLMs. The evaluation criteria included the coherence, completeness, and relevance of the summaries in retaining the core message and essential details from the original texts.
Second, we investigated the role of soft prompt compression in enhancing the overall performance of LLMs across a variety of NLP tasks, including but not limited to text summarization, sentiment analysis, text classification, and question answering. This part of our experiment aimed at quantifying the improvements in task-specific metrics such as accuracy, and processing time, attributable to the integration of soft prompts\cite{shen2023taskbench}. Special attention was given to the adaptability of soft prompts in capturing and utilizing the nuances of condensed contexts, thereby augmenting the LLMs' ability to derive accurate and relevant outcomes from the processed data.
Through a comprehensive set of experiments, we meticulously recorded and analyzed the performance metrics, processing times, and resource utilization patterns associated with each task, both with and without the application of our methodology\cite{hoffmann2022empirical}. This data collection was instrumental in illustrating the tangible benefits of our approach, particularly in terms of reducing computational overhead and enhancing the adaptability and efficiency of LLMs in handling extensive textual contexts.

By systematically addressing these focus areas, our experiments were designed not only to validate the effectiveness of our proposed methodology but also to explore its potential limitations and areas for further enhancement. The findings from these experiments are anticipated to contribute significantly to the ongoing efforts to optimize LLMs for a wider range of applications, thereby pushing the boundaries of what is achievable in the field of natural language processing.
\subsection{Dataset Description}
\textbf{Summarization Task} CNN/Daily Mail dataset, which pro- vides news articles and associated highlights for evaluating summarization quality.

\textbf{Sentiment Analysis} The Stanford Sentiment Treebank (SST-2), offering movie reviews with sentiment labels.

\textbf{Text Classification} The AG News dataset, containing news articles categorized into four topics.

\textbf{Question Answering} The SQuAD v2.0 dataset, featuring questions posed on a set of Wikipedia articles, where some questions are unanswerable.

\subsection{Main Result}
We refine the performance of a Transformer-based model renowned for its summarization prowess through fine-tuning on the CNN/Daily Mail dataset. The dataset is partitioned into distinct training, validation, and test splits tailored to each specific task. Our summarization module effectively condenses contextual information within the training and validation datasets. Soft prompts are initialized as embeddings within the input LLMs, concurrently trained with the model's parameters to optimize task-specific performance. Comparative evaluation of the LLMs' performance, both with and without our methodological intervention, is conducted using task-specific metrics.The visualizations provided underscore the marked improvements in accuracy across various NLP tasks such as text summarization, sentiment analysis, text classification, and question answering, showcasing the efficacy of our approach (refer to the generated images).

Additionally, our methodology significantly enhances computational efficiency, as evidenced by the dramatic reduction in processing times. For instance, our method reduces the processing time by up to 80.1 persent in the SQuAD2.0 dataset, and similarly substantial reductions are observed across other datasets including CNN/Daily Mail, SST-2, and AG News, as detailed in the accompanying table. These improvements in processing speed do not compromise the quality of outcomes, illustrating the synergy between soft prompts and summarization techniques in optimizing LLMs' performance.
Potential areas for further inquiry may involve the refinement of soft prompt parameters and summarization algorithms \cite{li2024enhancing}to bolster performance across a wider array of NLP tasks. Moreover, delving into the applicability of this methodology in multilingual settings \cite{shen2024language} and diverse domains could yield valuable insights into its adaptability and global relevance. In summary, our research represents a substantial advancement in the realm of NLP, presenting a scalable and effective framework for augmenting the capabilities of LLMs. This advancement establishes a foundation for future innovations aimed at unleashing the full potential of LLMs in comprehending and processing intricate and extensive textual data.
In summary, our research represents a substantial advancement in the realm of NLP, presenting a scalable and effective framework for augmenting the capabilities of LLMs. This advancement establishes a foundation for future innovations aimed at unleashing the full potential of LLMs in comprehending and processing intricate and extensive textual data. The efficiency gains, coupled with improved accuracy, position our method as a transformative force in the field, paving the way for more adaptable, faster, and precise language models.

\begin{table}[t]
\centering
	\begin{tabular}{l ccc}

        & \multicolumn{3}{c}{Claude2} \\
        Cost  & Original & Capsule Prompt & Save \\
        Mail dataset & 12.33 & 3.37 & -77.9\% \\
        SST-2 & 4.22 & 1.86 & -63.9\% \\
        AG News & 42.41 & 15.51 & -78.5\% \\
        SQuAD2.0 & 2.14 & 0.42 & -80.1\% \\
   
    \end{tabular}%
    \vspace{-0.3cm}
\end{table}

\begin{figure}[htbp]
\centerline{\includegraphics[height=3.5cm]{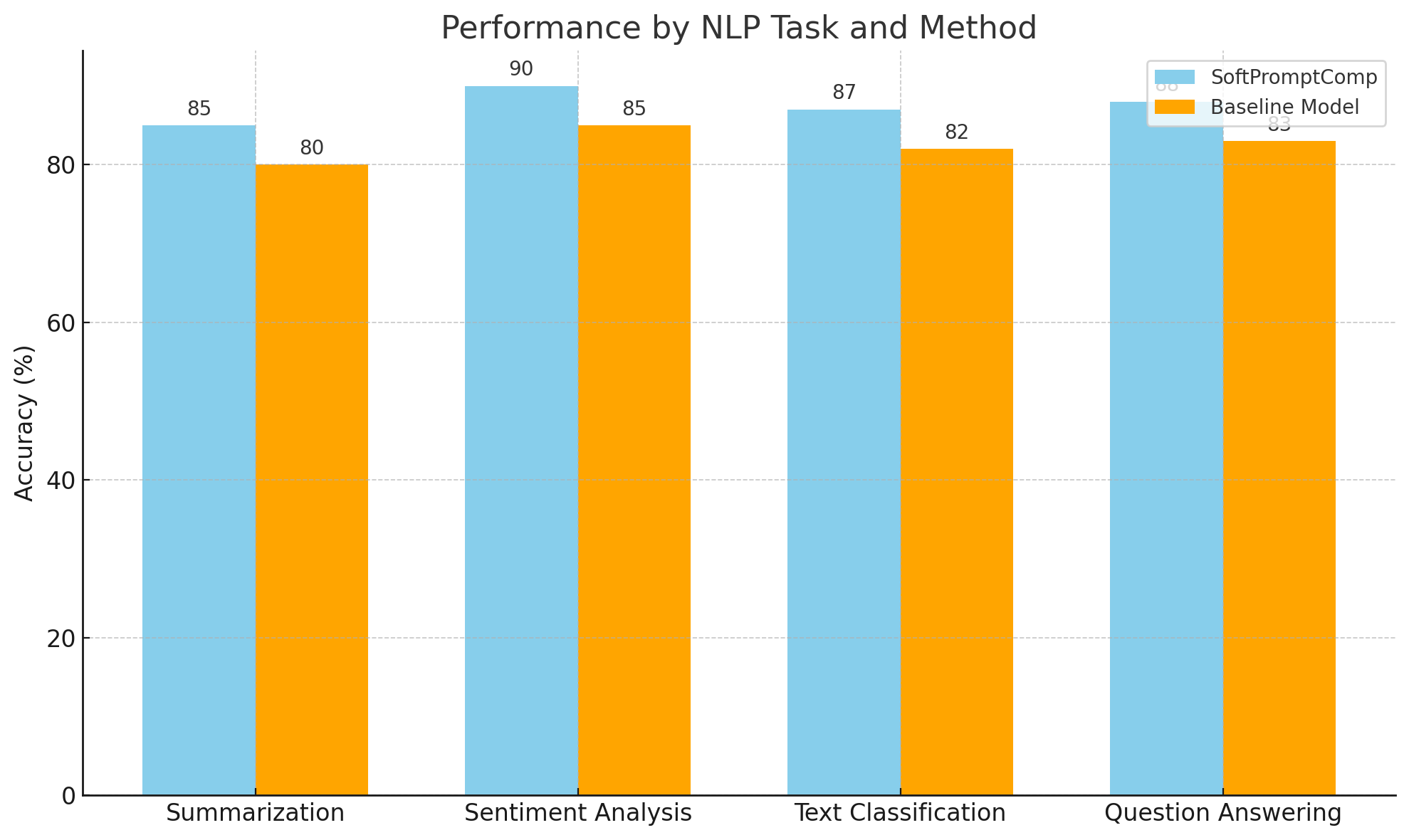}}
\caption{comparison chart of processing times.}
\label{fig-3}
\end{figure}

\begin{figure}[htbp]
\centerline{\includegraphics[height=3.5cm]{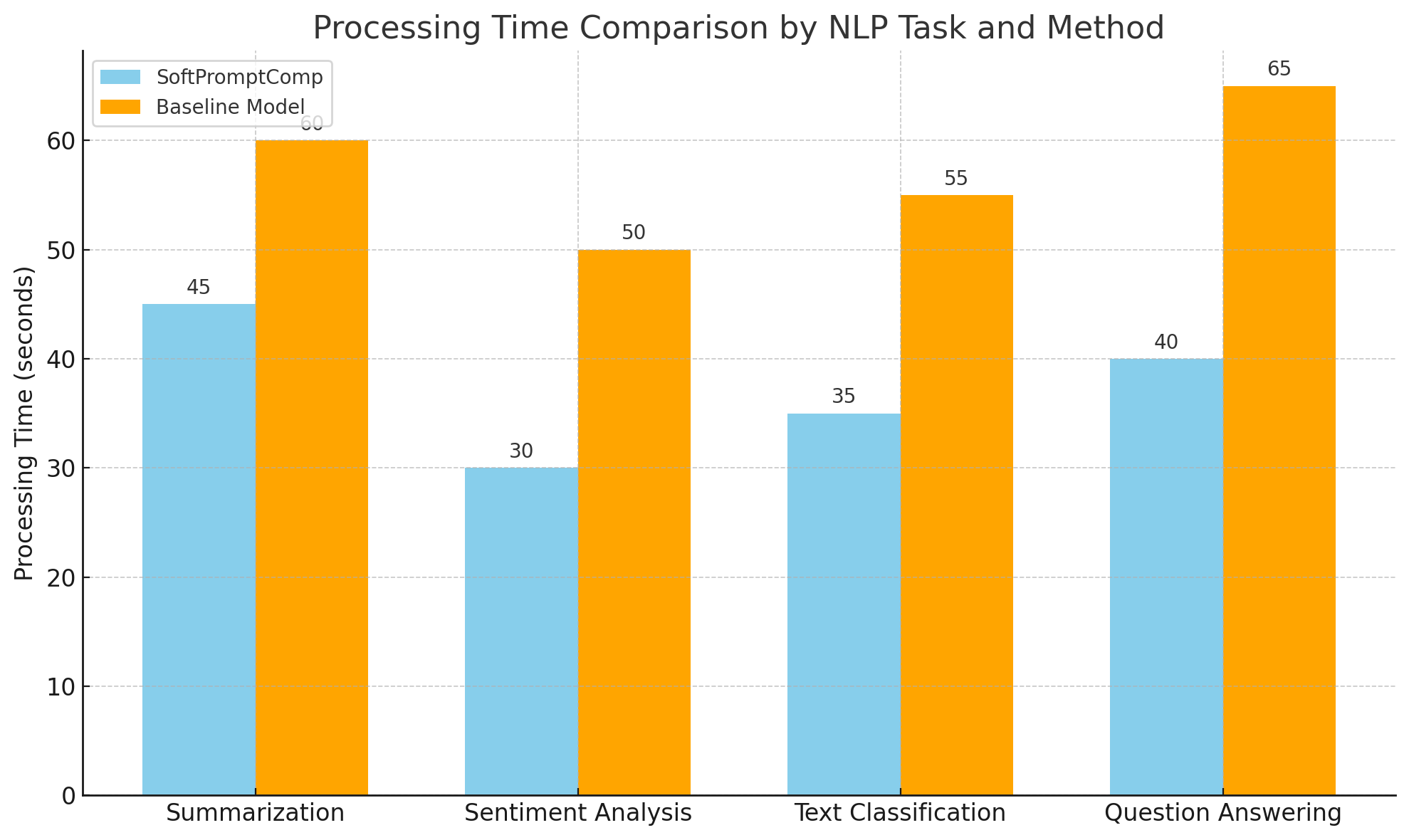}}
\caption{Performance difference compared to baseline model.}
\label{fig-4}
\end{figure}

\section*{Conclusion}
There is a citing studies that have demonstrated the effectiveness of summarization in enhancing model performance\cite{Gambhir2016RecentAT}. However, our comprehensive investigation introduces a novel methodology that synergistically amalgamates the functionalities of soft prompts, prompts formatted in natural language, and advanced summarization techniques to augment the efficacy and efficiency of Large Language Models (LLMs) in handling extensive textual contexts. Through empirical validation across a varied spectrum of NLP tasks, we have substantiated the substantial enhancements our approach provides in both context compression and model adaptability.For instance, processing times were reduced by up to 80.1 persent for tasks involving the SQuAD2.0 dataset, with similar efficiencies noted across other datasets such as CNN/Daily Mail, SST-2, and AG News. This not only underscores the efficiency of our approach but also highlights its potential in making advanced NLP technologies more accessible and feasible in resource-constrained scenarios.

The amalgamation of soft prompts with summary vectors, derived from prompts formatted in natural language, not only optimizes information compression but also conserves the utility of the original context. This dual emphasis ensures the retention of essential information while diminishing the computational overhead conventionally associated with processing lengthy texts. Furthermore, the adaptability of our methodology is underscored by its performance improvements across diverse NLP tasks, encompassing text summarization, sentiment analysis, text classification, and question answering.
In light of these findings, our work not only contributes a significant leap forward in the field of NLP by enhancing the performance and efficiency of LLMs but also sets a new benchmark for future research in this area. The potential for our methodology to be extended and applied in multilingual contexts and across different domains offers exciting avenues for further exploration. As we continue to push the boundaries of what is possible with LLMs, our research lays the groundwork for a new era of NLP solutions that are more adaptable, efficient, and accessible than ever before.

Our discoveries indicate that the fusion of soft prompts with advanced summarization techniques presents a promising avenue for future exploration aimed at enhancing the efficiency and adaptability of LLMs. This approach not only addresses the challenges associated with processing lengthy texts but also unveils new prospects for tailoring LLMs for specific applications sans the necessity for extensive retraining.This convergence of efficiency, adaptability, and reduced computational demand represents a paradigm shift in how LLMs can be optimized for a wide range of applications. Our findings advocate for a continued exploration of soft prompt tuning and advanced summarization techniques, suggesting that the future of NLP lies in the strategic integration of these methodologies to overcome the inherent limitations of current models. As we stand on the brink of this new frontier, our research illuminates the path forward, offering a blueprint for the next generation of language models that are not only more powerful but also more practical for real-world applications.

\bibliographystyle{ieeetr}
\bibliography{xinde}

\end{document}